\documentclass[sigconf]{acmart}
\usepackage{cleveref}
\usepackage{makecell}
\usepackage{multirow}
\usepackage{url}
\usepackage{booktabs}
\usepackage{subcaption}
\usepackage{pifont}
\usepackage{xcolor}
\usepackage{enumitem}
\usepackage{color}
\crefname{section}{Sec.}{Secs.}
\Crefname{section}{Section}{Sections}
\crefname{table}{Tab.}{Tabs.}
\Crefname{table}{Table}{Tables}
\crefname{figure}{Fig.}{Figs.}
\Crefname{figure}{Figure}{Figures}

\newcommand{\cmark}{\ding{51}} 
\newcommand{\xmark}{\ding{55}} 

\definecolor{stpipink}{rgb}{0.87,0.43,0.51}

\AtBeginDocument{%
  }
\renewcommand\footnotetextcopyrightpermission[1]{}
\begin{document}

\title{ST-$\pi$: Structured  SpatioTemporal  VLA  for  Robotic  Manipulation}

\author{Chuanhao Ma\textsuperscript{\rm 1}, Hanyu Zhou\textsuperscript{\rm 2}\footnotemark[1], Shihan Peng\textsuperscript{\rm 1}, Yan Li\textsuperscript{\rm 2}, Tao Gu\textsuperscript{\rm 3}, Luxin Yan\textsuperscript{\rm 1}\\
  \textsuperscript{\rm 1} School of Artificial Intelligence and Automation, Huazhong University of Science and Technology\\
  \textsuperscript{\rm 2} School of Computing, National University of Singapore\\
  \textsuperscript{\rm 3} School of Computing, Macquarie University\\
  {\tt\small {\{machuanhao, yanluxin\}}@hust.edu.cn, hy.zhou@nus.edu.sg}
}

\renewcommand{\shortauthors}{Ma et al.}

\begin{abstract}
Vision-language-action (VLA) models have achieved great success on general robotic tasks, but still face challenges in fine-grained spatiotemporal manipulation. Typically, existing methods mainly embed spatiotemporal knowledge into visual and action representations, and directly perform a cross-modal mapping for step-level action prediction. However, such spatiotemporal reasoning remains largely implicit, making it difficult to handle multiple sequential behaviors with explicit spatiotemporal boundaries. In this work, we propose \textbf{ST-$\pi$}, a structured spatiotemporal VLA model for robotic manipulation. Our model is guided by two key designs:
\textbf{1) Spatiotemporal VLM.} We encode 4D observations and task instructions into latent spaces, and feed them into the LLM to generate a sequence of causally ordered chunk-level action prompts consisting of sub-tasks, spatial grounding and temporal grounding.
\textbf{2) Spatiotemporal action expert.} Conditioned on chunk-level action prompts, we design a structured dual-generator guidance to jointly model spatial dependencies and temporal causality, thus predicting step-level action parameters.
Within this structured framework, the VLM explicitly plans global spatiotemporal behavior, and the action expert further refines local spatiotemporal control.
In addition, we propose a real-world robotic dataset with structured spatiotemporal annotations for fine-tuning. Extensive experiments have been conducted to demonstrate the effectiveness of our model. Our code link: \textcolor{stpipink}{\url{https://github.com/chuanhaoma/ST-pi}}.
\renewcommand{\thefootnote}{\fnsymbol{footnote}}
\footnotetext[1]{Corresponding author.}
\end{abstract}

\begin{teaserfigure}
  \includegraphics[width=\textwidth]{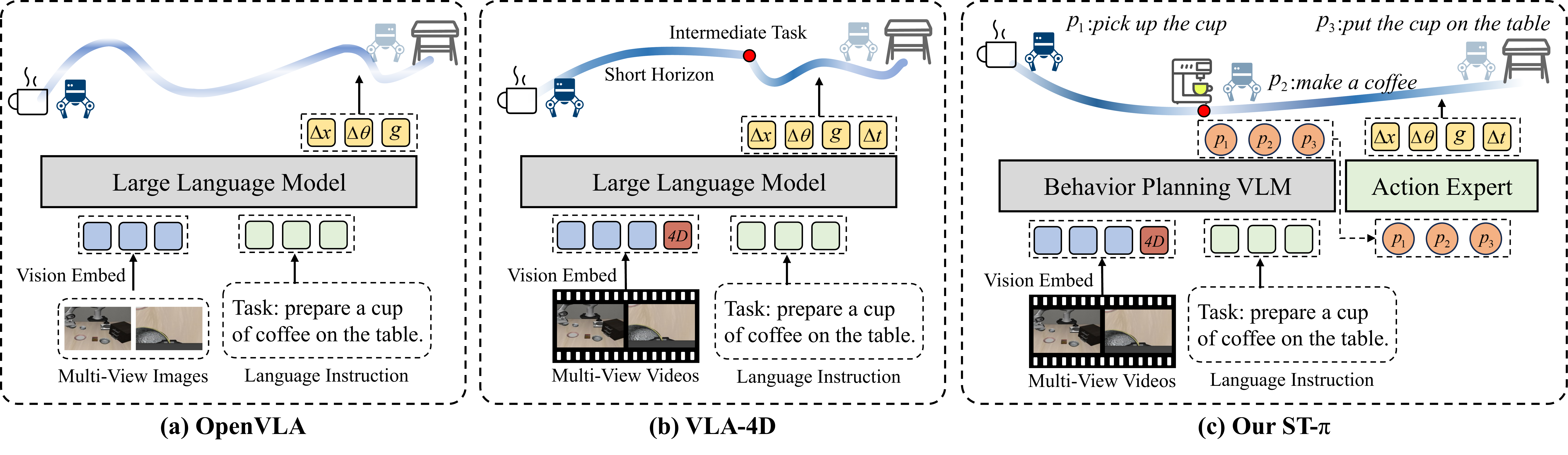}
  \caption{Illustration of typical VLA paradigms for spatiotemporal robotic manipulation.
  (a) OpenVLA operates on single-frame observations, leading to spatiotemporal chaos in long-horizon trajectories.
  (b) VLA-4D incorporates temporal information into visual and action representations, achieving consistent trajectories over short horizons but exhibiting instability in later stages.
  (c) Our ST-$\pi$ explicitly models structured task decomposition and employs a spatiotemporal action expert, enabling coherent and stable trajectories for long-horizon manipulation.
  }
  \Description{Comparison of different VLA paradigms for robotic manipulation.}
  \label{fig:teaser}
\end{teaserfigure}

\maketitle
\thispagestyle{plain}
\pagestyle{plain}

\begin{figure*}[t]
\centering
\includegraphics[width=\linewidth]{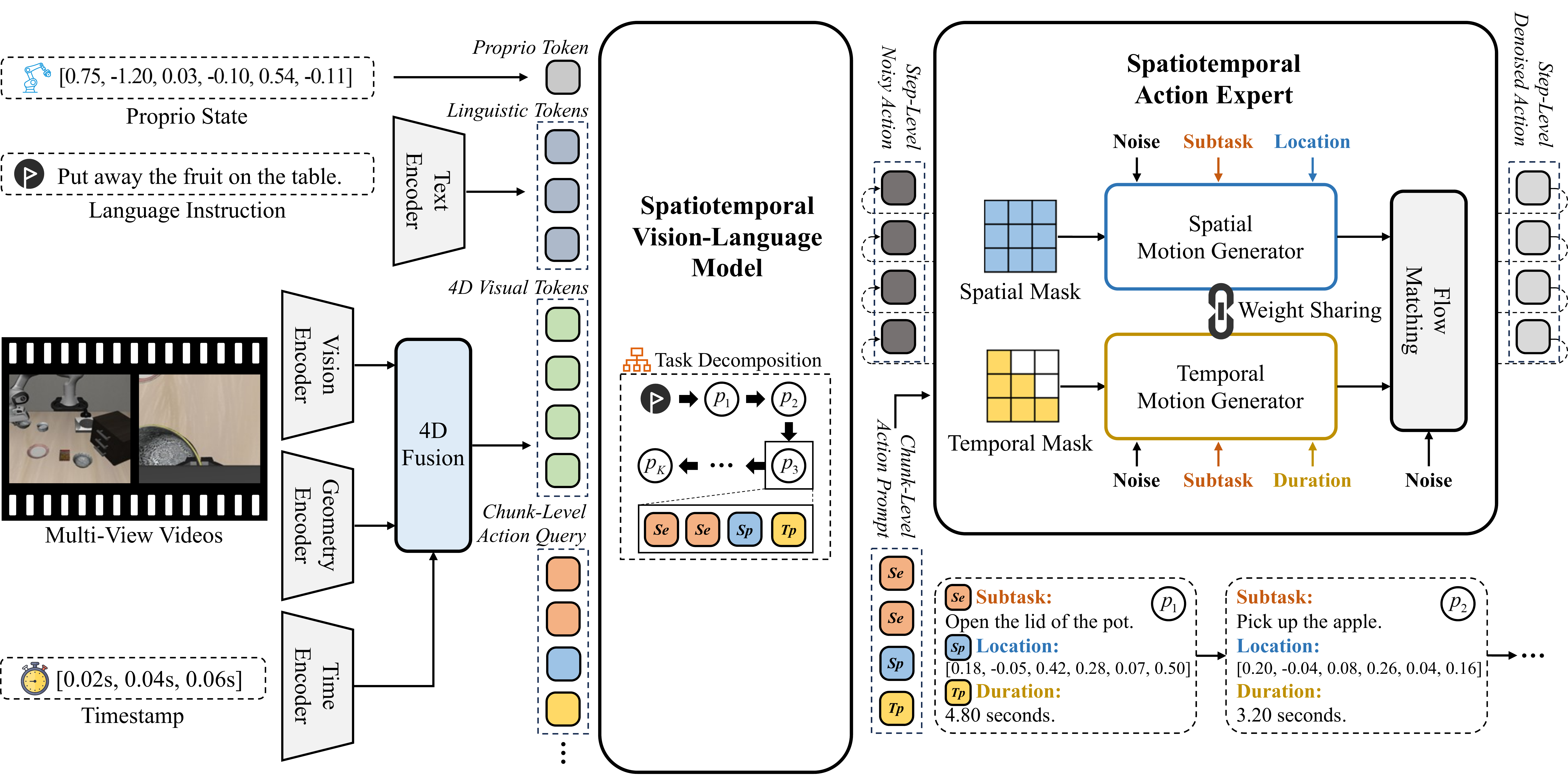}
\setlength{\abovecaptionskip}{-4pt}
\setlength{\belowcaptionskip}{0pt}
\caption{
Overview of ST-$\pi$.
Our model consists of two key components: (1) \textbf{ST-VLM}, which constructs 4D observation representations and performs structured task decomposition; and (2) \textbf{ST-AE}, which generates executable action chunks conditioned on chunk-level action prompts for each sub-task, ensuring stable and coherent trajectories.
}
\label{fig:framework}
\end{figure*}

\section{Introduction}
\label{sec:intro}
Vision-language-action (VLA) models~\cite{team2024octo,kim2024openvla,li2024cogact,black2024pi_0,li2025simplevla} have emerged as a promising paradigm for robotic manipulation that integrates perception, language reasoning, and action generation in a unified framework. Despite their strong performance on a wide range of robotic tasks, they still face significant challenges in fine-grained spatiotemporal manipulation tasks. Such tasks are typically long-horizon and composed of multiple sub-tasks with explicit spatial and temporal boundaries, such as complex assembly and household activities. Conventional VLA models~\cite{zheng2024tracevla,qu2025spatialvla} embed geometric information but lack explicit spatiotemporal representations, making it difficult to capture and reason about the structured boundaries in complex tasks. Therefore, our goal is to develop spatiotemporal models for fine-grained manipulation.

Existing 4D VLA approaches enhance spatiotemporal reasoning by extending spatial representations with temporal dimensions in both visual inputs and action representations. For instance, Zhang~\textit{et~al.}~\cite{4dvla} introduce a memory-bank-based historical frame-sampling strategy to construct 4D observation inputs; Zhou~\textit{et~al.}~\cite{vla4d} further extend action representations with temporal control to promote coherent action execution. However, since the spatiotemporal reasoning remains largely implicit, these methods are primarily limited to short-horizon behavior planning. While implicit representations are sufficient for short-horizon coherence, they fail to properly model the causal dependencies and transitions between sub-tasks in long-horizon scenarios. Therefore, it is essential to explicitly model spatiotemporal structure, enabling the representation of sub-task boundaries and reasoning about causal dependencies across long-horizon manipulation.

To address these issues, we argue that fine-grained manipulation requires explicit and structured spatiotemporal modeling at both chunk-level planning and step-level execution. For chunk-level planning, we suggest that complex tasks are composed of a sequence of sub-tasks with distinct spatial and temporal boundaries, where temporal causal dependencies exist across sub-tasks. This motivates us to explicitly model such dependencies and capture spatiotemporal boundaries for structured task decomposition. For step-level execution, we observe that action steps within the same sub-task exhibit inherent spatial coherence and temporal causality. This indicates the need to explicitly guide action generation along both spatial and temporal dimensions, ensuring spatial coherence and temporal stability. Consequently, a unified framework that explicitly models both chunk-level planning and step-level execution is essential for fine-grained spatiotemporal manipulation.

In this work, we propose \textbf{ST-$\pi$}, a structured spatiotemporal VLA framework for fine-grained robotic manipulation. As illustrated in \cref{fig:framework}, our ST-$\pi$ consists of a spatiotemporal vision-language model and a spatiotemporal action expert. For the spatiotemporal vision-language model, we construct 4D observations from image sequences and predict chunk-level action prompts containing semantic intent with spatial and temporal attributes. For the spatiotemporal action expert, we construct spatial and temporal motion generators to refine the actions and enforce spatiotemporal coherence and causality. In addition, we construct a real-world dataset, termed STAR, to support training, where manipulation trajectories are segmented into sub-tasks with semantic, spatial, and temporal annotations. The contributions of this work are as follows:
\begin{itemize}[leftmargin=10pt,itemsep=1pt]
\item We propose \textbf{ST-$\pi$}, a unified framework that explicitly structures both chunk-level task decomposition and step-level action generation, enabling fine-grained manipulation.
\item We design ST-VLM, a spatiotemporal VLM for behavior planning. The ST-VLM predicts causally ordered sub-tasks as chunk-level action prompts with semantic, spatial, and temporal attributes.
\item We introduce ST-AE, a structured action expert with complementary spatial and temporal generators, enabling spatially coherent and temporally consistent action generation.
\item We construct STAR, a real-world long-horizon robotic manipulation dataset with structured sub-task annotations. Extensive experiments demonstrate consistent improvements.
\end{itemize}
\begin{figure*}[t]
\centering
\includegraphics[width=\linewidth]{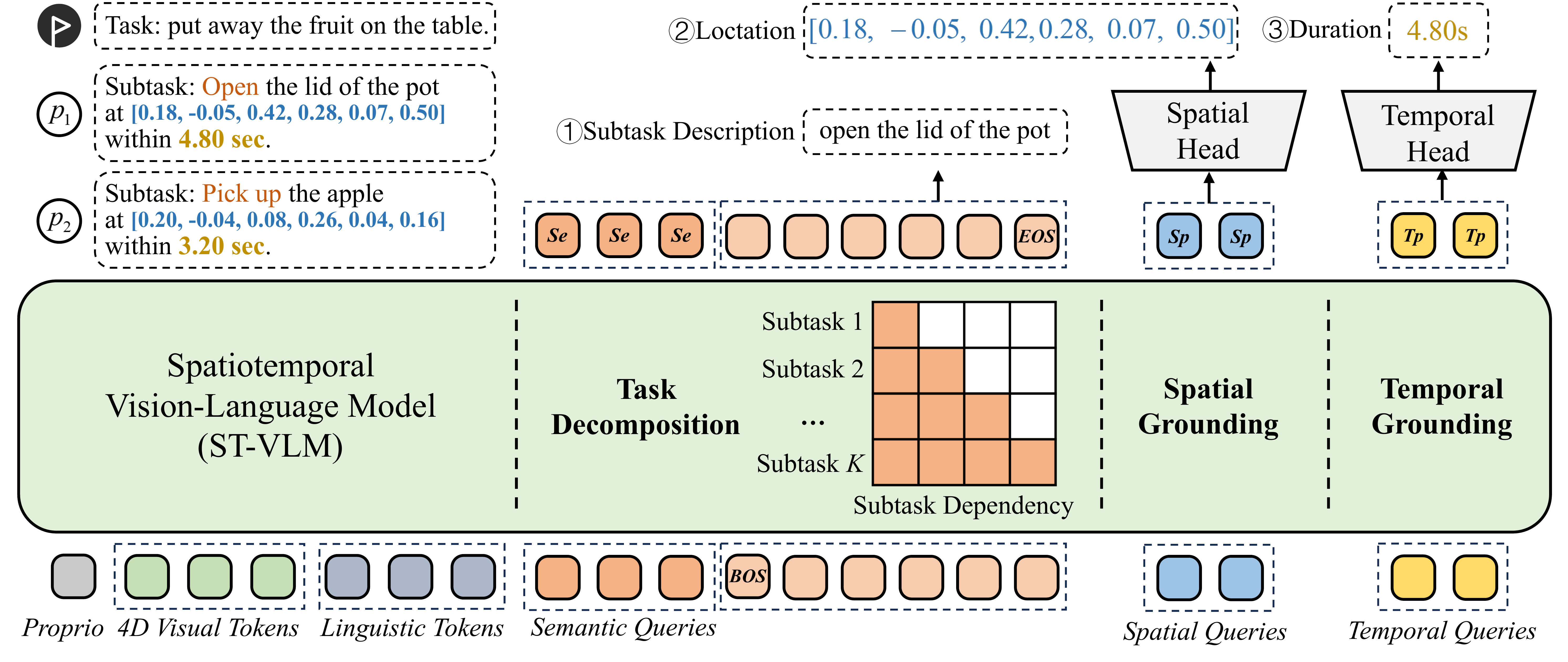}
\setlength{\abovecaptionskip}{-4pt}
\setlength{\belowcaptionskip}{0pt}
\caption{
Architecture of ST-VLM.
The ST-VLM takes 4D representations, language instructions, and query tokens as inputs to construct unified spatiotemporal representations. It performs structured task decomposition to generate chunk-level action prompts, including semantic intent and spatiotemporal information of sub-tasks.
}
\label{fig:stvlm}
\end{figure*}
\section{Related Works}
\noindent\textbf{Vision-Language-Action Models.}
The success of vision-language models (VLM)~\cite{liu2023visual, paligemma, bai2025qwen3} in scene reasoning promotes the adoption in robotics, giving rise to vision-language-action (VLA) models. Early VLA models~\cite{diffusionpolicy,team2024octo,kim2024openvla,zhou2025autovla} adopt end-to-end architectures that encode images, texts, and actions into a shared representation, directly mapping multimodal inputs to control parameters via cross-modal attention. These approaches benefit from large-scale multimodal pretraining~\cite{liu2023visual, hou20254d, zitkovich2023rt,brohan2022rt} and demonstrate strong generalization across diverse tasks. More recently, hierarchical VLA models~\cite{li2024cogact,pi05,black2024pi_0,lohovla,saycan} have been introduced, combining high-level reasoning with low-level control. However, most VLA frameworks~\cite{qu2025spatialvla,zhen20243d,li2025pointvla, zheng2024tracevla} primarily encode geometric cues while underutilizing temporal information. This motivates the exploration of 4D VLA frameworks integrating geometric knowledge with temporal information for stable and precise manipulation.

\vspace{3mm}\noindent\textbf{4D Vision-Language-Action Models.}
Recent works~\cite{niu2025pre, 4dvla, vla4d, hou20254d} have explored 4D VLA models to integrate temporal information with perception and control modeling. For example, Zhang~\textit{et~al.}~\cite{4dvla} augment visual inputs by performing geometric alignment and historical frame sampling to construct 4D observations, enhancing spatiotemporal perception from image sequences. However, temporal modeling is primarily applied at the perception level, while action dynamics remain implicitly modeled. Zhou~\textit{et~al.}~\cite{vla4d} further extend temporal modeling to the action space by embedding temporal information into both visual and action representations. Despite these advances, their spatiotemporal reasoning remains largely implicit. While effective for single-stage manipulation, the implicit formulation limits fine-grained spatiotemporal execution in multi-stage tasks. This inspires us to develop the spatiotemporal VLA model with explicit structured reasoning.

\vspace{3mm}\noindent\textbf{4D Reasoning Models.}
Recent advances in large multimodal models (LMMs)~\cite{llava4d, liu2023visual, 4drgpt, bai2025qwen3, zhou2025uni4d, wang2025vggt} significantly enhance the capability of spatiotemporal reasoning over dynamic scenes. These 4D LMMs extend visual-language modeling beyond static images with geometry cues and temporal dynamics, enabling temporally consistent grounding, cross-frame reasoning, and query-based understanding. For example, Yang~\textit{et~al.}~\cite{4drgpt} extend language models with temporally aligned region-level visual representations, enabling cross-frame grounding over dynamic scenes. Similarly, Zhou~\textit{et~al.}~\cite{llava4d} embed dynamic-aware 4D coordinate prompts into disentangled spatial and temporal visual features, aligning spatiotemporal embeddings with language representations to enhance 4D scene understanding. While these methods demonstrate strong capability, their spatiotemporal reasoning remains primarily perception-oriented rather than structured for embodied manipulation. This highlights the need for a structured spatiotemporal VLA framework explicitly decomposing high-level instructions into executable sub-tasks with spatiotemporally coherent actions.
\begin{figure*}[t]
\centering
\includegraphics[width=\linewidth]{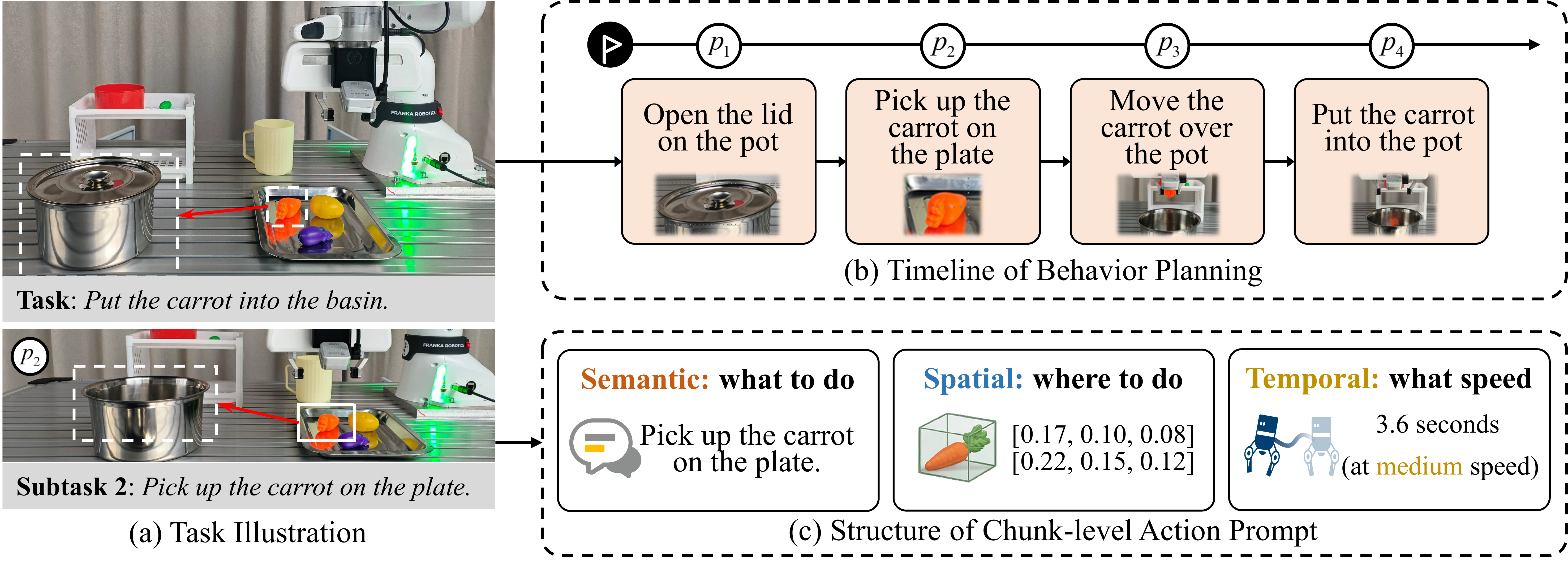}
\caption{
Overview of structured task decomposition with chunk-level action prompt.
A long-horizon task (a) is decomposed into a sequence of sub-tasks (b).
Each sub-task is represented as a chunk-level action prompt (c) with semantic, spatial, and temporal components, capturing what to do, where to act, and how to execute.
}
\label{fig:action_prompt}
\end{figure*}
\section{Our ST-$\pi$}
\noindent\textbf{Overview.} As illustrated in \cref{fig:framework}, our ST-$\pi$ consists of two key components: \textbf{1) SpatioTemporal Vision-Language Model} (ST-VLM) for chunk-level task decomposition, and \textbf{2) SpatioTemporal Action Expert} (ST-AE) for step-level action generation.

\vspace{1mm}\noindent\textbf{Our Framework.} Given a sequence of multi-view observations $\mathcal{I}_t = \{ I_{t-W+1}, \dots, I_t \}$ with $W$ frames, and the high-level language instruction $L$, we employ ST-VLM to decompose the task into a sequence of chunk-level action prompts, and then use ST-AE to generate executable action parameters conditioned on these prompts.

\begin{enumerate}[label={}, leftmargin=0pt]
\item \textbf{1) SpatioTemporal Vision-Language Model (\textit{cf.~\cref{ssec:stvlm}}).}
The ST-VLM constructs the 4D observation and decomposes high-level instructions. We first extract visual features $f_v$ and geometric features $f_g$ using a vision encoder and a geometry encoder, respectively. These features, together with timestamps $\mathbf{t}$, are then fused into unified 4D representations $f_{4D}$ via a spatiotemporal fusion module:
\begin{equation}
f_{4D}=\text{STE}(f_v, f_g, \mathbf{t}),
\end{equation}
where $\text{STE}(\cdot)$ denotes the spatiotemporal fusion module.
Conditioned on the 4D representations and the language instruction, the ST-VLM autoregressively predicts a sequence of chunk-level action prompts $\{p_k\}^{K}_{k=1}$ to represent sub-tasks, where $K$ denotes the maximum number of sub-tasks for task decomposition.
\item \textbf{2) SpatioTemporal Action Expert (\textit{cf.~\cref{ssec:stae}}).}
The ST-AE generates action parameters conditioned on the current sub-task $p_k$ via a flow-matching~\cite{flowmatching} process. We construct two complementary motion generators: a spatial generator and a temporal generator.
Conditioned on the action prompt $p_k$ and 4D representations, the two motion generators separately produce update flows $v_s$ and $v_t$, which are then fused via a flow fusion mechanism:
\begin{equation}
v_s=\text{SG}(p_k, f_{4D}), \quad v_t=\text{TG}(p_k, f_{4D}), \quad v=\text{FF}(v_s,v_t),
\end{equation}
where $v$ denotes the fused flow; $\text{SG}(\cdot)$ and $\text{TG}(\cdot)$ denote the spatial and the temporal generators; $\text{FF}(\cdot)$ denotes the flow fusion module. Then the fused flow guides the generation of action chunk $A_k$.
\end{enumerate}

\subsection{SpatioTemporal Vision-Language Model}
\label{ssec:stvlm}
As illustrated in \cref{fig:stvlm}, the ST-VLM performs structured chunk-level task decomposition conditioned on 4D representations and language instructions. Instead of implicitly encoding sub-tasks into latent features, ST-VLM represents them as chunk-level action prompts with semantic intent and spatiotemporal information, and explicitly models inter-chunk dependencies. In this section, we describe how ST-VLM constructs structured 4D representations and performs causal task decomposition.

\vspace{1mm}\noindent\textbf{Structured 4D Representation Construction.}
In robotic manipulation, conventional approaches typically rely on static 2D observations to predict action parameters in 3D space, leading to an inherent mismatch between perception and action. Extending observations to 4D representations improves the understanding of dynamic scenes. Therefore, we construct 4D representations by integrating visual features, geometric cues, and temporal contexts. We extract visual features using a SigLIP-based vision encoder~\cite{zhai2023sigmoid} and geometric features using a geometry encoder adapted from VGGT~\cite{wang2025vggt}. While visual features capture appearance semantics, geometric features encode spatial structure in latent space. To incorporate temporal contexts, we encode the timestamps $\mathbf{t}$ into temporal embeddings using Fourier positional encoding~\cite{li2021learnable} $\phi(\cdot)$. The temporal embedding $\phi(\mathbf{t})$ is fused with visual and geometric features as structured spatiotemporal embeddings via a linear layer:
\begin{equation}
f_v=\mathrm{VE}(\mathcal{I}_t), \quad f_g=\mathrm{GE}(\mathcal{I}_t), \quad f_{4D} = w_\mathrm{F}[f_v || f_g || \phi(\mathbf{t})],
\end{equation}
where $\mathrm{VE}(\cdot)$ and $\mathrm{GE}(\cdot)$ denote the vision and geometry encoders. This structured 4D representation provides explicit spatiotemporal grounding for the causal task decomposition in ST-VLM.

\vspace{1mm}\noindent\textbf{Structured Planning with Action Prompts.}
As illustrated in \cref{fig:action_prompt}, complex robotic manipulation tasks are typically composed of multiple sub-tasks, each with its own execution objective, operational region, and expected duration, while causal dependencies exist across sub-tasks. Our ST-VLM explicitly models such sub-task structures using chunk-level action prompts to capture the spatiotemporal information, and further enforces explicit constraints on the causal dependencies.
Given the 4D representation $f_{4D}$ and the high-level instruction $L$, the ST-VLM autoregressively predicts a sequence of action prompts within a short planning horizon $K$. Each prompt $p_k$ is composed of semantic tokens $s_k$, spatial tokens $\mathbf{x}_k$, and temporal tokens $\tau_k$:
\begin{equation}
p_k=\text{ST-VLM}(f_{4D}, L, p_{<k}),\quad p_k=\{s_k, \mathbf{x}_k, \tau_k\},
\end{equation}
where $p_{<k}$ denotes the preceding action prompts before step $k$. The ST-VLM follows a rolling-horizon scheme: ST-VLM predicts the next $K$ sub-tasks and re-plans after each sub-task is completed.

In each chunk-level action prompt, semantic tokens consist of $M$ learnable query tokens that capture the semantic intent and are supervised by the textual description of each sub-task, while the spatial and temporal tokens respectively encode the target execution region and the expected duration. Separate train-only regression heads are attached to tokens for supervision. During modeling, all tokens attend to the 4D representations to extract spatiotemporal context, while the spatial and temporal tokens further attend to the semantic tokens to explicitly model execution constraints.
Across action prompts, we model causal dependencies using a block-wise causal attention structure~\cite{vaswani2017attention,pi05}. The semantic tokens of prompt $p_k$ attend to all preceding prompts $p_{<k}$, while future prompts $p_{>k}$ are masked, enforcing causally ordered task decomposition.
\begin{figure}[t]
\centering
\includegraphics[width=\linewidth]{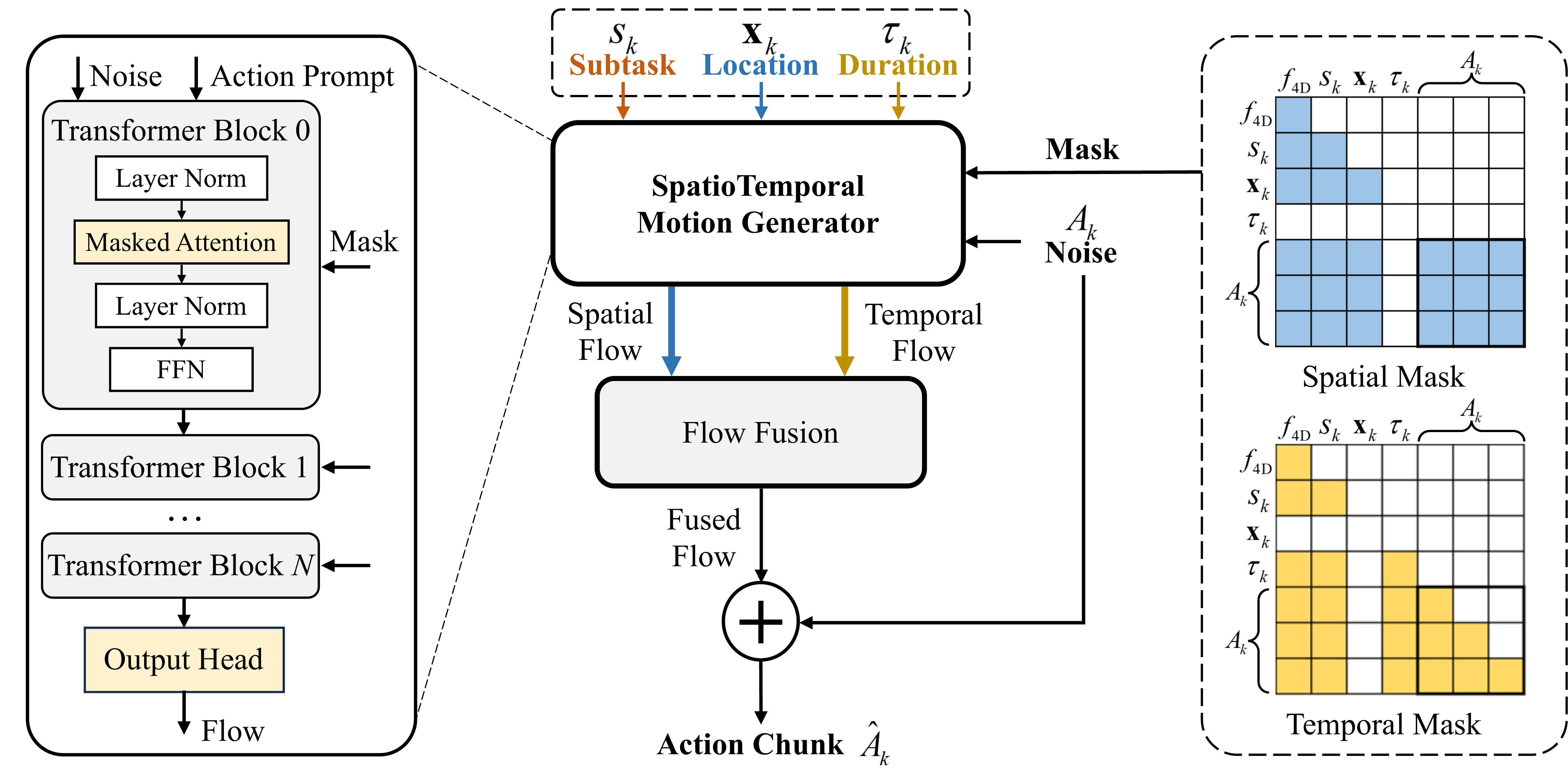}
\setlength{\abovecaptionskip}{-4pt}
\setlength{\belowcaptionskip}{0pt}
\caption{
Architecture of ST-AE.
The ST-AE takes chunk-level action prompts to generate actions through a dual-generator mechanism. A shared backbone with different attention constraints is applied to produce spatial and temporal flows, which are fused to guide action generation.
}
\label{fig:stae}
\end{figure}
\subsection{SpatioTemporal Action Expert}
\label{ssec:stae}
Conditioned on chunk-level action prompts, the ST-AE generates action chunks for each sub-task via a flow-matching process, illustrated in \cref{fig:stae}. Each action chunk $A_k$ consists of $H$ action steps:
\begin{equation}
A_k=\{a_k^{(i)}\}_{i=1}^{H}, \quad a_k^{(i)}=[\Delta\mathbf{x}, \Delta\mathbf{\theta}, g, \Delta t],
\end{equation}
where $\Delta\mathbf{x}$ and $\Delta \mathbf{\theta}$ denote translational and rotational motions of the robot; $g$ denotes the gripper motion; $\Delta t$ denotes the step duration.

\vspace{1mm}\noindent\textbf{Dual-Generator Spatiotemporal Guidance.}
Action generation involves two complementary stages: spatial shaping and temporal refinement. Spatial shaping transforms the initial noise into trajectory structures, while temporal refinement further adjusts the trajectory by enforcing causal dependencies across actions. We introduce a spatiotemporal dual-generator guidance: a spatial motion generator and a temporal motion generator, producing spatial update flows $v_s$ and temporal update flows $v_t$, respectively:
\begin{equation}
v_s=\Phi_s(f_{4D}, A_k, s_k, \mathbf{x}_k; \theta), \quad v_t=\Phi_t(f_{4D}, A_k, s_k, \tau_k; \theta),
\end{equation}
where $\Phi_s(\cdot)$ and $\Phi_t(\cdot)$ denote the spatial and temporal generators, and $\theta$ denotes their shared parameters.

The two generators differ in architectural constraints and conditioning features. The spatial motion generator conditions on the semantic and spatial tokens with a bidirectional attention constraint to action tokens. With all action steps in the chunk attending to each other, the spatial generator captures global spatial structures and encourages coordinated updates across steps to generate smooth trajectories. In contrast, the temporal motion generator conditions on the semantic and temporal tokens with a causal attention constraint to capture step-wise temporal dependencies and enforce sequential consistency. Each step is generated conditioned only on its preceding steps, ensuring temporally consistent action generation within the action chunk. Both motion generators are implemented using a shared backbone network to learn a unified action representation. This complementary design introduces distinct inductive biases, enabling the model to jointly capture spatial coordination and temporal causality for coherent action generation.

\vspace{1mm}\noindent\textbf{Dual-Generator Action Generation.}
In the early phase of action generation, spatial shaping plays a dominant role, establishing the coarse trajectory structure. In the later phase, temporal refinement becomes dominant, capturing causal dependencies and fine-tuning individual action steps. We adopt a time-dependent fusion strategy to gradually shift the guidance from spatial shaping to temporal refinement over the generation process:
\begin{equation}
v^{\tau} = \alpha^{\tau} v_t + (1-\alpha^{\tau}) v_s, \quad \alpha^{\tau} = \frac{\tau}{T},
\end{equation}
where $\tau$ denotes the current step; $T$ denotes the total number of denoising steps; $\alpha^\tau$ denotes the time-dependent fusion weight. Following the flow-matching formulation, the action chunk $A_k$ is refined by integrating the fused update flow:
\begin{equation}
\hat{A}_k=A_k+v^\tau\Delta \tau,
\end{equation}
where $\Delta \tau$ denotes the denoising step size. Starting from an initial Gaussian noise sample, the action chunk is iteratively refined, resulting in spatially coherent and temporally consistent trajectories.

\subsection{Optimization}
Our ST-$\pi$ involves two learning objectives: structured task planning for ST-VLM and dual-generator action generation for ST-AE.

\vspace{1mm}\noindent\textbf{Structured Task Planning.}
Given the 4D observation representation $f_{4D}$ and the high-level instruction $L$, we optimize the ST-VLM to predict a sequence of structured action prompts $\{p_k\}_{k=1}^K$. The training objective consists of three components: 1) Language modeling loss, which supervises the semantic tokens to predict the sub-task textual description. 2) Spatial regression loss, which supervises the spatial token to localize the target region. 3) Temporal regression loss, which supervises the temporal token to estimate the execution duration. The overall objective is formulated as:
\begin{equation}
\mathcal{L}_{\text{VLM}}=-\lambda_{L}\mathbb{E}[\log P(L_k|f_{4D}, L, s_k)]+\lambda_{s}|\mathbf{x}_k - \mathbf{x}_k^{*}|_1 + \lambda_{\tau}|\tau_k - \tau_k^{*}|_1,
\end{equation}
where $L_k$, $\mathbf{x}_k ^{*}$ and $\tau_k^{*}$ denote the ground-truth description, spatial target, and duration of sub-task; $\lambda_{L}$, $\lambda_s$ and $\lambda_{\tau}$ denote loss weights.

\vspace{1mm}\noindent\textbf{Dual-generator Action Generation.}
\label{sssec:duak_gen_action_gen}
To train the ST-AE, we adopt a flow-matching objective for action generation. Given a ground-truth action chunk $A_k^{*}$, the ST-AE learns to transport the noisy action toward the target trajectory. Following the flow-matching formulation~\cite{flowmatching,rectifiedflow,esser2024scaling}, we construct the noisy action by interpolating between the ground-truth action and Gaussian noise. For denoising timestep $\tau \in [0,1]$ and noise $\omega \sim \mathcal{N}(0,I)$, we construct the noisy action chunk $A_k^{\tau,\omega}$ and the target update flow $u_k^{\omega}$:
\begin{equation}
A_k^{\tau,\omega}=\tau A_k^{*} + (1-\tau)\omega, \quad u_k^{\omega}=A_k^{*}-\omega.
\end{equation}

The fused flow $v$ in the ST-AE is trained to match the target update flow with the following objective:
\begin{equation}
\mathcal{L}_{\text{AE}}=\mathbb{E}_{\tau,\omega,k}\left[\left\|v(A_k^{\tau, \omega})-u_k^\omega\right\|^2\right].
\end{equation}

\vspace{1mm}\noindent\textbf{Total Loss.}
\label{sssec:total_loss}
The entire framework is optimized end-to-end with a weighted sum of two objectives, which is formulated as:
\begin{equation}
\mathcal{L}=\lambda_{1}\mathcal{L}_{VLM}+\lambda_{2}\mathcal{L}_{AE},
\end{equation}
where $\lambda_{1}$ and $\lambda_{2}$ balance the two objectives.

\begin{figure*}[!t]
\centering
\includegraphics[width=\linewidth]{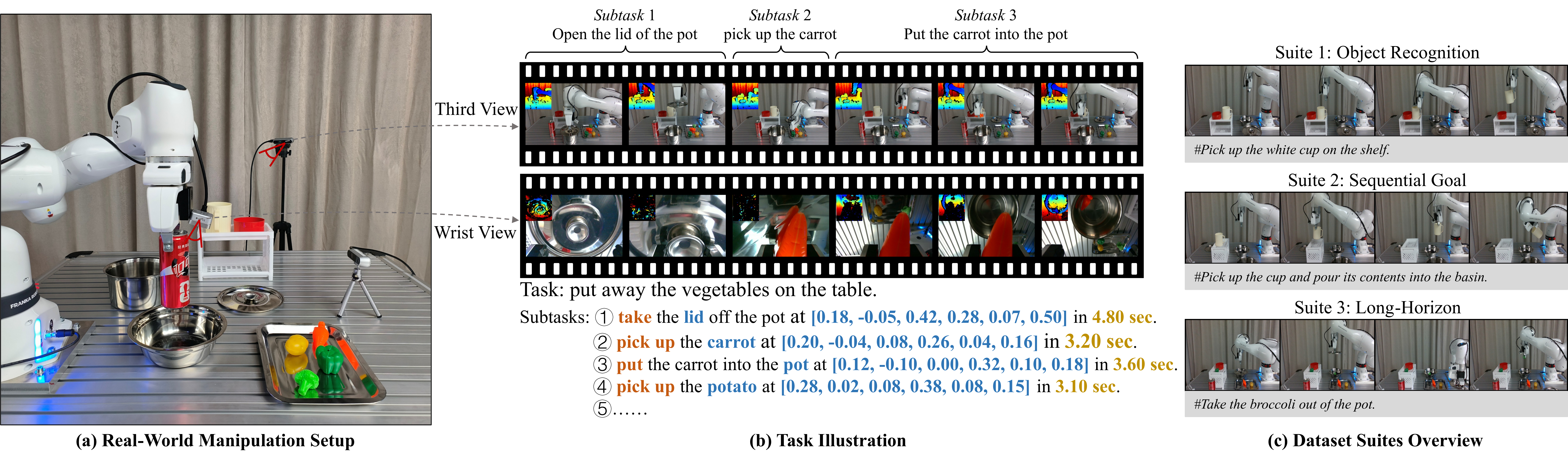}
\setlength{\abovecaptionskip}{-6pt}
\setlength{\belowcaptionskip}{-4pt}
\caption{
Illustration of our real-world robotic platform and STAR dataset.
Each task in the STAR dataset is segmented into multiple sub-tasks, with annotations including language instructions, target locations, and execution durations. The dataset includes three suites: Object Recognition, Sequential Goal, and Long-Horizon.
}
\label{fig:dataset}
\end{figure*}
\section{Training Datasets and Pipeline}
\subsection{Dataset}
To train and evaluate our ST-$\pi$, we leverage three complementary datasets to provide supervision at different levels.

\vspace{1mm}\noindent\textbf{ScanNet series.}
We utilize the ScanNet~\cite{scannet}, ScanRefer~\cite{chen2020scanrefer}, and Scan2Cap~\cite{chen2021scan2cap}, which provide richly annotated 3D reconstructions of indoor scenes. We extract RGB observations, descriptions, and corresponding 3D bounding boxes to provide spatial grounding supervision. These datasets are primarily used to train spatial representations and align geometric features with visual representations.

\vspace{1mm}\noindent\textbf{DROID-ST.}
\label{sssec:droid_st}
We construct DROID-ST, a structured task decomposition dataset derived from DROID~\cite{khazatsky2024droid} demonstrations. Each trajectory is decomposed into a variable-length sequence of sub-tasks, annotated with sub-task descriptions, target object bounding boxes, durations, and corresponding action segments.

\vspace{1mm}\noindent\textbf{STAR.}
\label{sssec:star}
As shown in \cref{fig:dataset}, we construct \textbf{STAR} (Spatiotemporal Task Annotation for Robotics), a real-world dataset including 30 manipulation tasks on the Franka Research 3 platform. The dataset is collected via a Gello~\cite{wu2023gello} teleoperation system with multi-view observations, where each task contains 50 demonstrations, resulting in around 300k interaction steps in total. The dataset consists of three suites: \textit{Object Recognition}, \textit{Sequential Goal}, and \textit{Long-Horizon}, covering a diverse set of tasks with increasing complexity. Each sub-task is annotated with descriptions, target locations, and execution durations. In terms of modalities, the inputs include multi-view images with depth observations, language instructions, and robot proprioceptive states. The outputs include end-effector motions and joint angles. To enhance temporal generalization, we collect demonstrations with varying speeds and encode the speed into the instruction using descriptors \textit{fast}, \textit{medium}, and \textit{slow}.

\subsection{Training Recipe}
Our ST-$\pi$ is trained through a three-stage training procedure that progressively enhances the capability of spatiotemporal perception, structured task planning, and robotic manipulation policy.

\vspace{1mm}\noindent\textbf{Stage 1: Representation Alignment.}
We first align geometric features with visual representations using ScanNet datasets. To preserve the pretrained knowledge of large backbones, we freeze the VLM backbone, the vision encoder and the geometry encoder, and train the geometry-aware adapters, the 4D fusion module and the spatial token with 3D bounding box supervision. In this stage, we only apply the spatial regression loss.

\vspace{1mm}\noindent\textbf{Stage 2: Behavior Planning Learning.}
In the second stage, we train the model to perform structured task decomposition using the DROID-ST dataset. First, we freeze the time encoder and temporal token, and train the ST-VLM along with semantic tokens. Conditioned on the 4D representation and the language instruction, the model is supervised to predict the textual description of each sub-task, with the language modeling loss and LoRA~\cite{hu2022lora} applied for efficient adaptation. Next, we unfreeze and train the temporal components. The temporal token is supervised to predict the duration of each sub-task with the temporal regression loss, enabling the ST-VLM to capture temporal dynamics in task decomposition.

\vspace{1mm}\noindent\textbf{Stage 3: Robotic Task Fine-tuning.}
In the third stage, we perform end-to-end fine-tuning using the DROID-ST dataset to learn a unified manipulation policy. We freeze the vision encoder, the geometry encoder, the 4D fusion module and the action prompt tokens to preserve the decomposition capability learned previously. The model is then fine-tuned to generate executable action chunks conditioned on the 4D representation and the chunk-level action prompt, supervised by the flow-matching loss in \cref{sssec:duak_gen_action_gen}.

\begin{table*}[t]
\setlength{\abovecaptionskip}{1pt}
\setlength{\belowcaptionskip}{1pt}
\centering
\caption{Evaluation on the LIBERO benchmark.}
\label{tab:libero}
\begin{tabular}{lcccccccccc}
\Xhline{1pt}
\multicolumn{1}{c}{\multirow{2}{*}{Model}} & \multicolumn{2}{c}{Spatial} & \multicolumn{2}{c}{Object} & \multicolumn{2}{c}{Goal} & \multicolumn{2}{c}{Long} & \multicolumn{2}{c}{Average} \\
& SR $\uparrow$ & CT(s) $\downarrow$ & SR $\uparrow$ & CT(s) $\downarrow$ & SR $\uparrow$ & CT(s) $\downarrow$ & SR $\uparrow$ & CT(s) $\downarrow$ & SR $\uparrow$ & CT(s) $\downarrow$ \\
\hline
OpenVLA~\cite{kim2024openvla} & 84.7\% & 5.5 & 88.4\% & 7.5 & 79.2\% & 6.1 & 53.7\% & 13.1 & 76.5\% & 8.0 \\
Octo~\cite{team2024octo} & 78.9\% & 5.7 & 85.7\% & 6.9 & 84.6\% & 6.3 & 51.1\% & 9.3 & 75.1\% & 7.0 \\
SpatialVLA~\cite{qu2025spatialvla} & 88.2\% & 5.3 & 89.9\% & 7.4 & 79.0\% & 6.0 & 55.5\% & 8.9 & 78.1\% & 6.6 \\
TraceVLA~\cite{zheng2024tracevla} & 84.6\% & - & 85.2\% & - & 76.3\% & - & 54.1\% & - & 74.8\% & - \\
4D-VLA~\cite{4dvla} & 88.9\% & - & 95.2\% & - & 90.9\% & - & 79.1\% & - & 88.6\% & - \\
CogACT~\cite{li2024cogact} & 87.5\% & 5.4 & 90.2\% & 7.1 & 80.2\% & 6.0 & 53.2\% & 10.7 & 77.8\% & 7.2 \\
$\pi_{0.5}$~\cite{pi05} & \textbf{98.8\%} & \textbf{5.2} & 98.2\% & 6.3 & 96.8\% & 5.8 & 92.4\% & 7.8 & 96.9\% & 6.3 \\
ST-$\pi$(Ours) & 98.4\% & 5.4 & \textbf{98.3\%} & \textbf{6.0} & \textbf{96.9\%} & \textbf{5.3} & \textbf{94.3\%} & \textbf{6.9} & \textbf{97.3\%} & \textbf{5.9} \\
\Xhline{1pt}
\end{tabular}
\end{table*}
\begin{table*}[t]
\setlength{\abovecaptionskip}{1pt}
\setlength{\belowcaptionskip}{1pt}
\centering
\caption{Evaluation on the SIMPLER benchmark.}
\begin{tabular}{clcccccccc}
\Xhline{1pt}
\multicolumn{1}{c}{\multirow{2}{*}{Protocol}} & \multicolumn{1}{c}{\multirow{2}{*}{Model}} & \multicolumn{2}{c}{\#Pick Coke Can} & \multicolumn{2}{c}{\#Move Near} & \multicolumn{2}{c}{\#Open/Close Drawer} & \multicolumn{2}{c}{Average} \\
& & SR $\uparrow$ & CT(s) $\downarrow$ & SR $\uparrow$ & CT(s) $\downarrow$ & SR $\uparrow$ & CT(s) $\downarrow$ & SR $\uparrow$ & CT(s) $\downarrow$ \\
\hline
\multicolumn{1}{c}{\multirow{6}{*}{Visual Matching}} & OpenVLA~\cite{kim2024openvla} & 14.7\% & 15.2 & 49.2\% & 11.8 & 39.8\% & 15.5 & 34.6\% & 14.2 \\
& Octo~\cite{team2024octo} & 14.7\% & 11.8 & 4.2\% & 10.3 & 24.1\% & 18.4 & 14.3\% & 13.5 \\
& CogACT~\cite{li2024cogact} & \textbf{91.0\%} & 6.1 & 78.0\% & 6.4 & 75.9\% & 13.7 & \textbf{81.6\%} & 8.7 \\
& SpatialVLA~\cite{qu2025spatialvla} & 81.7\% & 6.2 & 83.3\% & 6.7 & 57.4\% & 14.8 & 74.1\% & 9.2 \\
& $\pi_{0.5}$~\cite{pi05} & 83.1\% & 6.1 & 81.4\% & 6.2 & 60.9\% & 14.4 & 75.1\% & 8.9 \\
& ST-$\pi$(Ours) & 85.5\% & \textbf{6.0} & \textbf{84.1\%} & \textbf{6.0} & \textbf{68.2\%} & \textbf{13.7} & 79.3\% & \textbf{8.6} \\
\hline
\multicolumn{1}{c}{\multirow{6}{*}{Variant Aggregation}} & OpenVLA~\cite{kim2024openvla} & 48.2\% & 12.6 & 43.5\% & 11.7 & 15.1\% & 17.6 & 35.6\% & 14.0 \\
& Octo~\cite{team2024octo} & 0.2\% & 15.8 & 2.8\% & 13.6 & 0.8\% & 25.8 & 1.3\% & 18.4 \\
& CogACT~\cite{li2024cogact} & \textbf{89.2\%} & 6.3 & 80.8\% & 6.6 & 25.8\% & \textbf{13.1} & 65.3\% & 8.7 \\
& SpatialVLA~\cite{qu2025spatialvla} & 85.7\% & 7.0 & 76.1\% & 7.5 & 25.9\% & 15.2 & 62.6\% & 9.9 \\
& $\pi_{0.5}$~\cite{pi05} & 84.6\% & 6.1 & 78.3\% & 6.8 & 31.2\% & 13.5 & 64.7\% & 8.8 \\
& ST-$\pi$(Ours) & 86.7\% & \textbf{5.8} & \textbf{81.4\%} & \textbf{6.2} & \textbf{33.6\%} & 13.3 & \textbf{67.2\%} & \textbf{8.4} \\
\Xhline{1pt}
\end{tabular}
\label{tab:simpler}
\end{table*}

\section{Experiments}
\subsection{Experiment Setup}
\noindent\textbf{Implementation Details.}
Our model adopts the pretrained weights of PaliGemma~\cite{paligemma} from $\pi_{0.5}$~\cite{pi05} as the VLM backbone, DINOv2~\cite{oquab2023dinov2} from VGGT~\cite{wang2025vggt} as the geometry encoder, and Gemma-300M~\cite{team2024gemma} equipped with a structured spatiotemporal attention mechanism as the action expert. The model is trained on 8 NVIDIA RTX PRO 6000 GPUs, using the AdamW~\cite{loshchilov2017decoupled} optimizer. In training stage 1, we set $\lambda_s=1$, $\lambda_{L}=\lambda_{\tau}=0$, and the learning rate to $2e-5$ with a batch size of 64. In stage 2 and stage 3, we set $\lambda_{L}=\lambda_{1}=1$, $\lambda_s=\lambda_{\tau}=5$, $\lambda_{2}=10$, and the learning rate to $1e-5$ with a batch size of 32.

\vspace{1mm}\noindent\textbf{Comparison Methods.}
We compare our model with several representative VLA approaches, including OpenVLA~\cite{kim2024openvla}, Octo~\cite{team2024octo}, Diffusion Policy~\cite{diffusionpolicy}, SpatialVLA~\cite{qu2025spatialvla}, TraceVLA~\cite{zheng2024tracevla}, 4D-VLA~\cite{4dvla}, $\pi_{0.5}$~\cite{pi05}, and CogACT~\cite{li2024cogact}. The first three methods are 2D VLA models operating on single-frame images. The next two methods are 3D VLA models incorporating 3D positions. 4D-VLA is the 4D method with spatiotemporal visual observations. The last two methods adopt dual-system architectures within the VLA framework.

\vspace{1mm}\noindent\textbf{Evaluation Metrics.}
We evaluate all models on three benchmarks: LIBERO~\cite{libero}, SIMPLER~\cite{simplerenv} and our \textbf{STAR}. For LIBERO and SIMPLER, models are trained on the corresponding dataset and evaluated within the same environment. For real-world experiments, we fine-tune models on the \textbf{STAR} dataset. We report task success rate and completion time as the primary evaluation metrics.

\begin{figure*}[t]
\centering
\includegraphics[width=\linewidth]{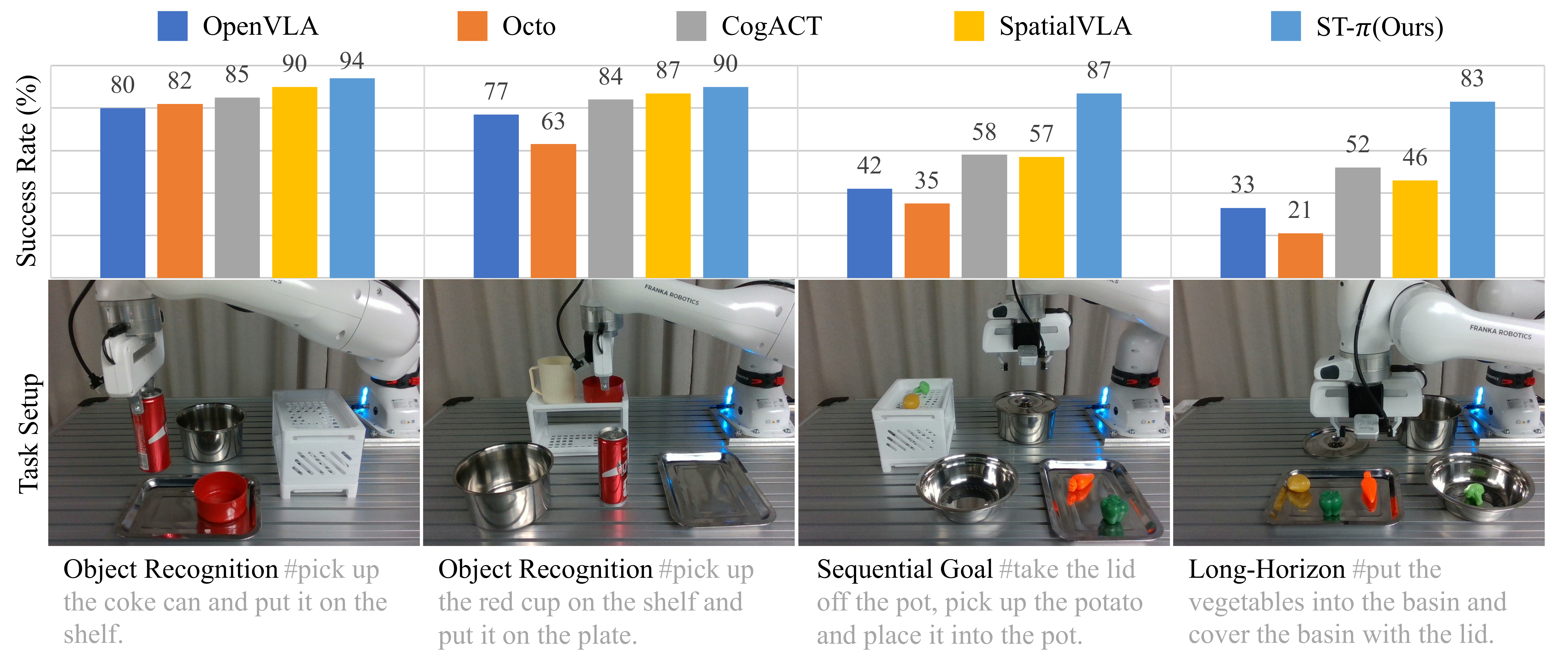}
\setlength{\abovecaptionskip}{-10pt}
\setlength{\belowcaptionskip}{-4pt}
\caption{
Real-world performance comparison.
We evaluate different VLA methods on real-world tasks with increasing complexity. Our ST-$\pi$ consistently outperforms all baselines, demonstrating robustness in long-horizon manipulation.
}
\label{fig:realworld}
\end{figure*}
\begin{table*}[t]
\centering
\setlength{\abovecaptionskip}{1pt}
\setlength{\belowcaptionskip}{1pt}
\caption{Evaluation on the real-world STAR benchmark.}
\begin{tabular}{lcccc}
\Xhline{1pt}
\multicolumn{1}{c}{Model} & Object Recognition & Sequential Goal & Long-Horizon & Average \\
\hline
OpenVLA~\cite{kim2024openvla} & 78.4\% & 47.0\% & 35.6\% & 53.7\% \\
CogACT~\cite{li2024cogact} & 81.2\% & 63.4\% & 52.4\% & 65.7\% \\
SpatialVLA~\cite{qu2025spatialvla} & 86.8\% & 59.6\% & 49.2\% & 65.2\% \\
$\pi_{0.5}$~\cite{pi05} & 89.2\% & 70.0\% & 69.4\% & 76.2\% \\
ST-$\pi$(Ours) & \textbf{92.0\%} & \textbf{75.8\%} & \textbf{72.8\%} & \textbf{80.1\%} \\
\Xhline{1pt}
\end{tabular}
\label{tab:real_world}
\end{table*}
\subsection{Comparison on Simulation Benchmarks}
\noindent\textbf{Evaluation on LIBERO.}
We fine-tune and evaluate all competing models in four suites of LIBERO. From the results in \cref{tab:libero}, we derive three main observations.
First, performance improves as observations become more expressive, evolving from 2D single-frame inputs to 3D spatial representations and further to 4D spatiotemporal observations.
Second, VLA models adopting dual-system architectures with high-level behavior planning and low-level action generation consistently outperform single-system counterparts.
Third, our model achieves the highest success rates and the shortest completion time on most tasks. This advantage is primarily attributed to the explicit modeling of spatiotemporal sub-task dependencies.
The results highlight the importance of structured spatiotemporal modeling in long-horizon manipulation tasks.

\vspace{1mm}\noindent\textbf{Evaluation on SIMPLER.}
We adopt SIMPLER~\cite{simplerenv} as an additional simulation benchmark for scalable evaluation. SIMPLER provides a suite of simulation environments that replicate common real-world robotic setups and supports real-to-sim evaluation. The benchmark consists of two evaluation protocols, \emph{Visual Matching} and \emph{Variant Aggregation}, which assess generalization under visual distribution shifts and environment variations. We compare our ST-$\pi$ with several representative baselines, including OpenVLA~\cite{kim2024openvla}, Octo~\cite{team2024octo}, CogACT~\cite{li2024cogact}, SpatialVLA~\cite{qu2025spatialvla}, and $\pi_{0.5}$~\cite{pi05}. As shown in \cref{tab:simpler}, our method achieves competitive performance across evaluation protocols, demonstrating efficient execution in simulations.

\subsection{Comparison on Real-World STAR}
To validate the practical applicability of our framework, we conduct real-world experiments on a Franka Research 3 platform. All competing methods are fine-tuned on the \textbf{STAR} dataset in \cref{sssec:star} and evaluated on three evaluation suites. As shown in \cref{fig:realworld} and \cref{tab:real_world}, our model achieves the highest average success rate across all evaluation suites. The performance gap becomes more pronounced in long-horizon manipulation scenarios, where precise behavior transitions and task decomposition are critical for planning. This result indicates that explicit structured spatiotemporal modeling of task decomposition enhances execution stability and robustness in real-world robotic manipulation environments.

\subsection{Ablation Study}
\noindent\textbf{Effect of the Structured Spatiotemporal Framework.}
To evaluate the effect of the structured spatiotemporal framework, we conduct an ablation study isolating the contributions of the ST-VLM and the ST-AE. We compare four variants by enabling or disabling (i) structured task decomposition in ST-VLM and (ii) spatiotemporal motion generators in ST-AE. When decomposition is disabled, the model directly generates actions for the entire task without predicting intermediate sub-tasks; when the structured ST-AE is removed, motion generators are replaced with a standard full-attention policy. As shown in \cref{tab:aba_framework}, both components contribute to performance. ST-VLM improves long-horizon manipulation stability by modeling chunk-level dependencies, while the ST-AE enhances action consistency by enforcing spatial coherence and temporal causality. The combination yields the best performance, demonstrating that structured planning and action generation are complementary and critical for fine-grained spatiotemporal manipulation.

\vspace{2mm}\noindent\textbf{Ablation on High-Level Spatiotemporal Attention Structure.}
We further analyze the effect of the spatiotemporal attention structure in ST-VLM for sub-task decomposition. In our ST-$\pi$, the predicted sub-tasks follow a causal attention constraint enforcing temporal ordering during decomposition. We compare three attention configurations in \cref{tab:aba_high_level_attn_mask}. Results show that the causal attention structure consistently yields the best performance across different task suites and evaluation metrics. Bidirectional attention weakens the temporal structure in decomposition, while removing inter-chunk attention entirely drops the dependencies between sub-tasks. This suggests that explicitly modeling causal dependencies is necessary for stable planning and coherent long-horizon execution.

\vspace{1mm}\noindent\textbf{Ablation on the Observation Modality.}
We further study the impact of the observation modality. Our full model leverages multi-frame observations with geometric features, forming structured 4D representations. We progressively simplify the observations and compare three settings: 4D observations (multi-frame images + geometric features), 3D observations (single-frame image + geometric features), and 2D observations (single-frame image only). Results in \cref{tab:aba_modality} show that performance consistently degrades as observations are simplified. Using multi-frame observations leads to more stable long-horizon execution, while geometric features provide spatial cues for precise manipulation. The full 4D representation achieves the best performance, demonstrating the importance of jointly modeling appearance, geometry, and temporal context.

\begin{figure*}[t]
\centering
\includegraphics[width=\linewidth]{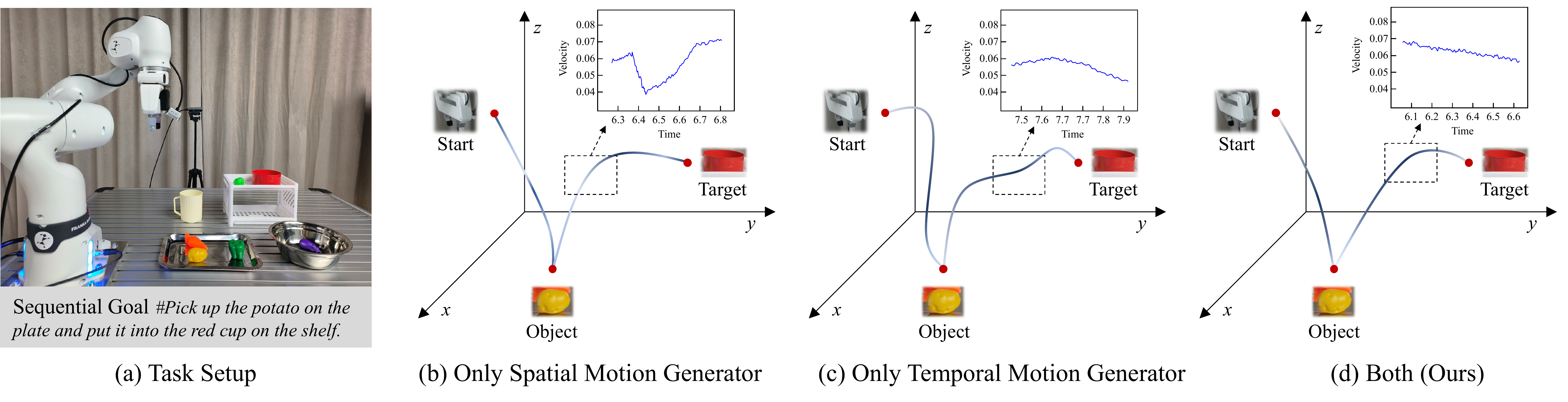}
\caption{
Trajectory comparison of different motion generator variants.
(a) Task illustration.
(b) Only the spatial motion generator produces smooth but irregular-velocity trajectories.
(c) Only the temporal motion generator produces nearly constant-velocity but non-smooth trajectories.
(d) Combining both yields smooth and temporally consistent trajectories.
}
\label{fig:3dtraj}
\end{figure*}

\begin{table}[t]
\setlength{\abovecaptionskip}{2pt}
\setlength\tabcolsep{3pt}
\setlength{\belowcaptionskip}{2pt}
\centering
\caption{Effect of Structured Spatiotemporal Framework.}
\renewcommand\arraystretch{1.05}
\begin{tabular}{cc|ccccc}
\Xhline{1pt}
\multirow{2}{*}{\makecell{SpatioTemporal\\VLM}} & \multirow{2}{*}{\makecell{SpatioTemporal\\Action Expert}} & \multicolumn{2}{c}{LIBERO Avg.} & \multicolumn{2}{c}{STAR} \\
& & SR$\uparrow$ & CT$\downarrow$ & SR$\uparrow$ & CT$\downarrow$ \\
\hline
\xmark & \xmark & 96.9\% & 6.3 & 76.8\% & 16.2 \\
\cmark & \xmark & 97.2\% & 6.2 & 78.4\% & 15.4 \\
\xmark & \cmark & 97.1\% & 6.0 & 77.2\% & 14.6 \\
\cmark & \cmark & \textbf{97.4\%} & \textbf{5.9} & \textbf{80.1\%} & \textbf{13.5} \\
\Xhline{1pt}
\end{tabular}
\label{tab:aba_framework}
\end{table}
\begin{table}[t]
\setlength{\abovecaptionskip}{2pt}
\setlength\tabcolsep{5pt}
\setlength{\belowcaptionskip}{2pt}
\centering
\caption{Ablation on High-Level Spatiotemporal Attention.}
\renewcommand\arraystretch{1.05}
\begin{tabular}{c|cccc}
\Xhline{1pt}
\multirow{2}{*}{Mask} & \multicolumn{2}{c}{LIBERO Avg.} & \multicolumn{2}{c}{STAR} \\
& SR$\uparrow$ & CT$\downarrow$ & SR$\uparrow$ & CT$\downarrow$ \\
\hline
None & 92.5\% & 6.1 & 73.8\% & 15.8 \\
Bidirectional & 96.5\% & 6.0 & 78.4\% & 14.5 \\
Causal (Ours) & \textbf{97.4}\% & \textbf{5.9} & \textbf{80.1}\% & \textbf{13.5} \\
\Xhline{1pt}
\end{tabular}
\label{tab:aba_high_level_attn_mask}
\end{table}

\subsection{Discussion}
\noindent\textbf{Role of Spatiotemporal Generators.}
We analyze how the spatiotemporal motion generators in ST-AE influence action generation behavior. In \cref{fig:3dtraj}, we visualize the trajectories under three configurations, each isolating the spatial and temporal generators. The spatial-only variant produces coherent trajectories, but exhibits irregular step sizes, leading to unstable temporal progression. In contrast, the temporal-only variant generates trajectories with consistent velocities, but lacks spatial smoothness, resulting in jagged motion. Our full model, which integrates both generators, produces trajectories that are both spatially smooth and temporally stable. This demonstrates that the spatial and temporal generators provide complementary inductive biases, and their joint modeling enables coherent and stable action generation.

\vspace{1mm}\noindent\textbf{Effect of Task Decomposition Granularity.} 
We analyze how the granularity of task decomposition influences performance by varying the number of predicted sub-tasks $K$. We evaluate $K\in\{1,2,3,4\}$ while keeping other settings fixed. Results in \cref{tab:dis_subtask_num} show that performance improves as the decomposition becomes more refined, indicating that moderate task decomposition promotes better structure of long-horizon behaviors. When $K$ increases to 4, performance slightly decreases compared to $K=3$, suggesting that overly fine-grained decomposition may introduce additional prediction noise or reflect insufficient training for deeper decomposition levels. These results indicate that a moderate decomposition granularity provides a trade-off between planning flexibility and stability.

\begin{table}[t]
\setlength{\abovecaptionskip}{2pt}
\setlength\tabcolsep{5pt}
\setlength{\belowcaptionskip}{2pt}
\centering
\caption{Ablation on Observation Modality.}
\renewcommand\arraystretch{1.05}
\begin{tabular}{c|cccc}
\Xhline{1pt}
\multirow{2}{*}{Modality} & \multicolumn{2}{c}{LIBERO Avg.} & \multicolumn{2}{c}{STAR} \\
& SR$\uparrow$ & CT$\downarrow$ & SR$\uparrow$ & CT$\downarrow$ \\
\hline
2D & 94.6\% & 6.4 & 74.2\% & 16.3 \\
3D & 96.0\% & 6.2 & 77.1\% & 15.2 \\
4D (Ours) & \textbf{97.4\%} & \textbf{5.9} & \textbf{80.1\%} & \textbf{13.5} \\
\Xhline{1pt}
\end{tabular}
\label{tab:aba_modality}
\end{table}

\begin{table}[t]
\setlength{\abovecaptionskip}{2pt}
\setlength\tabcolsep{5pt}
\setlength{\belowcaptionskip}{2pt}
\centering
\caption{Effect of Behavioral Decomposition Granularity.}
\renewcommand\arraystretch{1.05}
\begin{tabular}{c|cccc}
\Xhline{1pt}
\multirow{2}{*}{Subtask Number $K$} & \multicolumn{2}{c}{LIBERO Avg.} & \multicolumn{2}{c}{STAR} \\
& SR$\uparrow$ & CT$\downarrow$ & SR$\uparrow$ & CT$\downarrow$ \\
\hline
$K=1$ & 96.5\% & 6.7 & 76.2\% & 16.6 \\
$K=2$ & 97.1\% & 6.1 & 79.3\% & 14.3 \\
$K=3$ & \textbf{97.4}\% & \textbf{5.9} & \textbf{80.1}\% & 13.5 \\
$K=4$ & 97.2\% & 6.1 & 79.9\% & \textbf{13.4} \\
\Xhline{1pt}
\end{tabular}
\label{tab:dis_subtask_num}
\end{table}

\vspace{1mm}\noindent\textbf{Limitation.}
Despite its effectiveness, our framework assumes that robotic manipulation tasks can be decomposed into a sequential chain of sub-tasks. While this assumption holds for many robotic manipulation scenarios, more complex behaviors may involve parallel or branching structures that cannot be fully captured by a strictly causal decomposition. Extending the framework to support more flexible task structures and richer dependency modeling across sub-tasks remains a valuable direction for future work.

\section{Conclusion}
In this paper, we present ST-$\pi$, a structured spatiotemporal VLA framework for fine-grained robotic manipulation. The proposed framework introduces a spatiotemporal vision-language model that explicitly decomposes high-level instructions into structured chunk-level action prompts with semantic, spatial, and temporal representations. We further design a spatiotemporal action expert that generates step-level action parameters through a dual-generator guidance, ensuring spatial coherence and temporal causality during motion generation. To support structured supervision, we construct a real-world dataset termed STAR, with structured annotations for fine-grained spatiotemporal manipulation training. Extensive experiments on simulation benchmarks and real-world robotic platforms demonstrate that our framework achieves consistent improvements over representative VLA baselines.

\bibliographystyle{ACM-Reference-Format}
\bibliography{main}

\end{document}